# Locality-Aware Rotated Ship Detection in High-Resolution Remote Sensing Imagery Based on Multi-Scale Convolutional Network

Lingyi Liu, Yunpeng Bai, and Ying Li

*Abstract*—Ship detection has been an active and vital topic in the field of remote sensing for a decade, but it is still a challenging problem due to the large scale variations, the high aspect ratios, the intensive arrangement, and the background clutter disturbance. In this letter, we propose a locality-aware rotated ship detection (LARSD) framework based on a multi-scale convolutional neural network (CNN) to tackle these issues. The proposed framework applies a UNet-like multi-scale CNN to generate multi-scale feature maps with high-level semantic information in high resolution. Then, a rotated anchor-based regression is applied for directly predicting the probability, the edge distances, and the angle of ships. Finally, a locality-aware score alignment is proposed to fix the mismatch between classification results and location results caused by the independence of each subnet. Furthermore, to enlarge the datasets of ship detection, we build a new high-resolution ship detection (HRSD) dataset, where 2499 images and 9269 instances were collected from Google Earth with different resolutions. Experiments based on public dataset HRSC2016 and our HRSD dataset demonstrate that our detection method achieves state-of-the-art performance.

## I. INTRODUCTION

SHIP detection has been a topic of interest in the field of remote sensing over the last decades and has made great progress in promoting national defense construction, harbor management, and cargo transportation. With the rapid growth of satellite technology and construction, high-resolution optical remote sensing images can be easily obtained, which contain abundant details for classifying objects. Attributing to the advancement, ship detection from optical remote sensing images is under more active research. Fukun Bi et al. [1] employed the bottom-up visual attention mechanism and top-down visual cues for candidate selection and discrimination, respectively. Yang Guang et al. [2] integrated sea surface analysis into ship detection to make the detection method robust to the variation of sea surfaces. These ship detection methods implement ship detection by extracting and recognizing the features of shapes and textures from ships. Nonetheless, it is tricky to design the handcrafted features applied to all categories of ship.

Recently, deep learning algorithms, especially the convolutional neural network (CNN) have led to significant progress in object detection. CNN-based detection methods mainly include one-stage methods and two-stage methods. One-stage methods [3-5] regard object localization as a regression model, directly estimating the class and region of objects. Compared with one-stage methods, two-stage methods have higher accuracy but lower interference speed. Two-stage methods [6], [7] can be SEPARATED into two components. The region proposal stage generates proposal regions by a selected search algorithm or a CNN model, and the detection stage predicts the class probability of these generated regions and the offsets between ground-truths and these regions. These state-of-the-art object detectors for natural images have already achieved promising results. However, several challenges limit the application of these detectors in ship detection:

1) The sizes of ships range widely in remote sensing images due to the variety of ship categories and space resolutions, as shown in Fig. 1(a). The fixed single receptive field determined by the architecture of CNN models cannot match the scale variability of ships.
2) Ships are in stripe-like shapes and are often docked inshore or side-by-side intensively, as shown in Fig. 1(b). Using horizontal bounding boxes for location may cover a relatively large redundancy complex background region. Besides, for horizontal bounding boxes, it is also a challenging issue to distinguish the adjacent ships.
3) The remote sensing images captured by satellites generally have large size and contains numerous ship-like object (e.g. containers and fish rafts). These background clutter disturbances easily lead to false alarms on the detectors.
4) The existing ship detection datasets in optical satellite images are scarce, especially for rotated ship detection datasets.

Inspired by the success from natural images, the deep learning-based ship detection methods on optical remote sensing images have attracted more and more attention. For example, G. Cheng et al. [10] proposed a rotation-invariant CNN model that introduced a new rotation-invariant layer to address the problem of object rotation variations. Zhang, R. Q et al. [11] proposed a ship region proposal method combined with line segment features and saliency features. These methods above have already achieved encouraging performance in the field of remote sensing detection. However, the limitation of horizontal bounding boxes mentioned above poses a great challenge for inshore and adjacent ship detection.

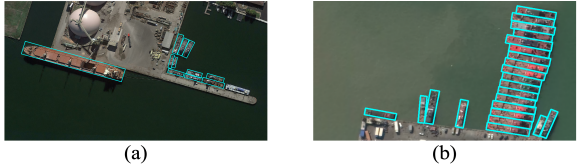

Fig. 1. Visualization of ships. (a) Illustration of different scale ships. (b) Illustration of intensively docked ships.

Several rotated ship detectors have also been proposed to overcome this limitation. Zikun Liu et al. [12] proposed a rotated region-based CNN for ship detection, which uses a rotated region of interest (RRoI) pooling layer to extract rotation region features and directly regress rotation angles. Xue Yang et al. [13] further proposed multi-scale rotation dense feature pyramid networks. They design a rotation anchor strategy with multi-scale RoI Align to improve the efficiency of the feature extract model for the rotated object. Nevertheless, large numbers of rotation anchors increase the difficulty of object classification and generate more false alarms. Mingjie Li et al. [14] proposed a rotated ship detector based on a fully convolutional network. This method has both accuracy and speed performance but displays low accuracy in the complex background because of the imbalance proportion of positive samples and negative samples. Minghui Liao et al. [15] proposed a rotation sensitive regression detector (RDD) which implements classification and regression using rotation-insensitive features and sensitive features, respectively. RDD improves the performance in case of a dense arrangement, but the extractor of rotation-insensitive features and sensitive features need to be enhanced for the recall of ship detection. Ding Jian et al. [16] proposed the RoI Transformer to effectively extract features by spatial transformations. It is an effective strategy for converting rotated RoI into horizontal RoI, but the operations of spatial transformations may lead to the confusion of features.

To date, several ship detection datasets have been released for research reference. Wang Y et al. [8] constructed the SAR ship detection dataset (SSDD), which is a ship detection dataset of SAR images and includes 1160 images and 2456 instances in total. The labels of SSDD are annotated with horizontal bounding boxes (x, y, w, h). Zikun Liu et al. [9] built a ship detection dataset in optical satellite images named high-resolution ship collection 2016 (HRSC2016). It contains 1061 images and 2886 samples collected from Google Earth. The size of images ranges from 300x300 to 1500x900, and the space resolutions are in the range of 0.4m and 2m. It is annotated with rotated bounding boxes. In general, these datasets promote the development of ship detection, but the limited number of public datasets restricts the further improvement of ship detection methods.

In this paper, we propose a one-stage rotated object detection framework based on multi-scale CNN. Specifically, our model first learns a UNet-like multi-scale CNN to extract multi-level feature maps. Then, an anchor-based rotated bounding box regression is utilized to generate some candidate targets. Moreover, the scores of the candidate targets are refined by locality-aware score alignment (LASA). Finally, we adopt non-maximum suppression to merge overlapping bounding boxes.

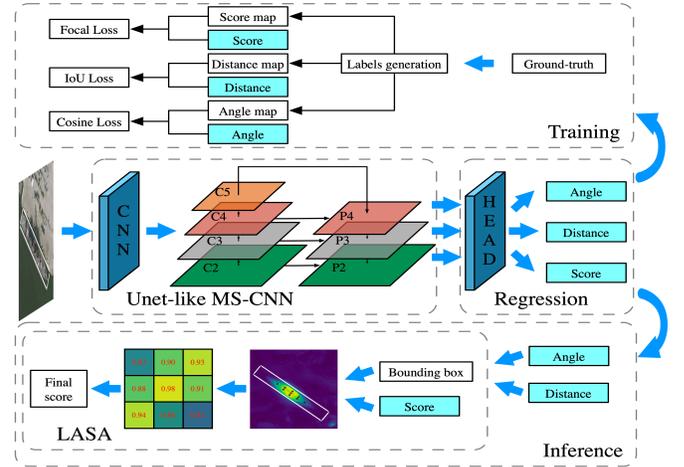

Fig. 2. Flowchart of Locality-Aware Rotated Ship Detection Based on Multi-Scale Convolutional Network.

The main contributions of this paper are as follow:
1) We built a novel ship detection framework based on a one-stage detection method and utilized an additional angle regression branch to predict rotated bounding boxes as object targets. Experimental results demonstrate that our proposed method outperforms state-of-the-art methods.
2) We proposed locality-aware score alignment to refine the scores of bounding boxes according to their locations, which alleviate the problem of mismatch between classification results and location results.
3) We introduced a new ship dataset labeled by oriented bounding boxes, which has 2499 images and 9269 instances. The dataset contains various ships with different scenarios, scales and space resolutions to eliminate the dataset bias.

## II. PROPOSED METHOD

In this section, we provide details for the proposed detection framework, which is comprised of three main components: UNet-like multi-scale CNN for feature extraction, anchor-based rotated bounding box regression for candidate target prediction, and locality-aware score alignment (LASA) for class score refinement. Fig. 2 shows the overall framework of our method.

### A. UNet-Like Multi-Scale CNN

Extracting semantic features efficiently from high-resolution remote sense images is crucial to the deep learning-based ship detection methods. Appropriate semantic features can distinguish ships from the complex background. ResNet-101 achieves promising performance in classification, yet the sizes of final feature maps are only 1/16 of original images. Although final feature maps have high-level semantic information, it loses more detailed information during several pooling operations, especially location information, which further increases the difficulty for small or high aspect ratio object detection such as ships. Inspired by the semantic segment task that requires both high-level semantic information and high resolution, which is similar to the demand for ship detection, we introduce the network structure of UNet. As shown in Fig. 3, {C2, C3, C4, C5} donates the feature maps of ResNet-101, feature map by nearest neighbor upsampling and concatenate the upsampling feature map with corresponded lower level feature map. A 3×3 convolution and a 1×1 convolution are used to fuse the feature maps and reduce the number of feature map channels. Undergoing three iterations, we merge high-level semantic information into the low-level feature map then expend the resolution of feature maps gradually. To recognize various scales of ships, we utilize multi-level feature maps as the input of rotated anchor-based regression and classification networks. The final set of feature maps is defined as {*P2, P3, P4*}, and we append a 1×1 convolution to equalize the channel numbers of the feature maps.

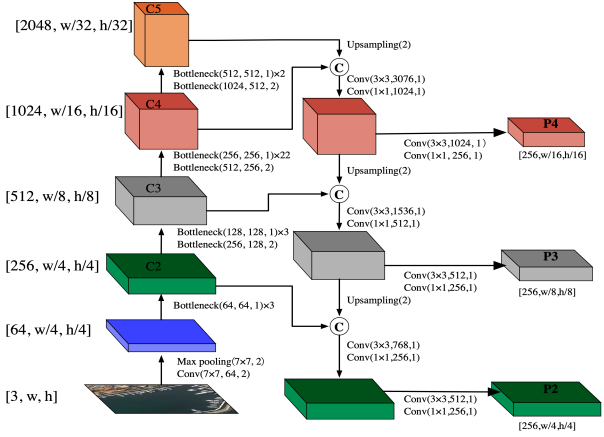

Fig. 3. The architecture of UNet-like multi-scale CNN.

*B. Rotated Bounding Box Regression*

As mentioned above, large scale variations and high aspect ratios are the main challenges for detecting ship location. Therefore, we adopt the anchor-based regression method and multi-scale prediction. Our prediction module includes two task-specific subnetworks: the classification subnet and the location regression subnet. The classification subnet predicts the probability of object for each anchor, while the location regression subnet predicts the distances from each position to the four sides of the bounding boxes.

*1) Rotated Bounding Boxes*

We use the five variables ($d_1, d_2, d_3, d_4, \theta$) to uniquely determine the rotated bounding box. As shown in Fig. 4, *x* and *y* are the coordinates of the anchor point in input images. {$D_1, D_2, D_3, D_4$} donates the corner points of rectangles. We set the point with the lowest sum of x and y as the $D_1$, and $D_2, D_3, D_4$ are as follows by clockwise. {$d_1, d_2, d_3, d_4$} are the distances from the anchor point to four sides of the rectangle. $\theta$ is the horizontal angle of the rotated bounding box, and we convert the range of $\theta$ from [-90, 0) to (-45, 45] for normalization.

*2) Anchor-Based Regression*

We use the anchor strategy to facilitate the regression of distance. The size of anchor priors is determined by k-means clustering. We divide the ground-truth into three groups according to their areas and choose five clusters at each group as the anchor priors.

Each anchor at every pixel in feature maps is assigned to a one-hot vector for anchor classification, a 5-dimensional vector for distance and angle regression. The assignment rule is based on the combination of intersection-over-union (IoU) and the range of rotated bounding boxes. Specifically, we first compute the IoU between ground-truth and anchor priors with the same center. Then, a location is set as a positive sample if it is in the range of ground-truth and its corresponding IoU is greater than 0.5, as an ignored sample if it is in the range of ground-truth and its IoU is in [0.2, 0.5], and as a background sample in other conditions.

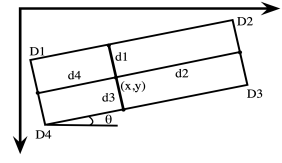

Fig. 4. Illustration of rotated bounding box.

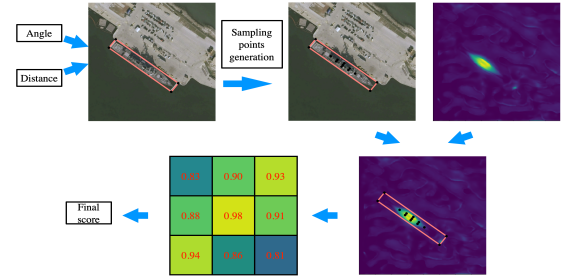

Fig. 5. Flowchart of the locality-aware score alignment.

The classification subnet applies three 3×3 convolution layers and a 1×1 convolution layer in sequence, followed by a sigmoid function as the normalized function. The output for each location in feature maps is a K-dimensional vector which predicts the scores of all K anchors. The location regression subnet also consists of three 3×3 convolution layers and a 1×1 convolution layer and export a 5-dimensional vector ($t_1, t_2, t_3, t_4, t_\theta$) for distance and angle regression. The distance and the angle are calculated in the formulas as below:

$$d_i = \begin{cases} h_k \exp(t_i), & \text{for } i = 1, 3 \\ w_k \exp(t_i), & \text{for } i = 2, 4 \end{cases} \quad (1)$$

$$\theta = (\text{sigmoid}(t_\theta) \times 2 - 1) \times \pi / 4$$

where $h_k$ and $w_k$ are the length and width of the *k*-th anchor.

*3) Loss Function*

Our loss function is represented below:

$$L = \frac{1}{N_{cls}} \sum_i L_{cls}(p_i, p_i^*) + \frac{1}{N_{reg}} \sum_i (p_i^* (\lambda_d L_d(\boldsymbol{d}_i, \boldsymbol{d}_i^*) + \lambda_a L_a(\theta_i, \theta_i^*))) \quad (2)$$

Where $p_i^*$ refers to the label of the object, $p_i$ is the predicted classification score, $\boldsymbol{d}_i$ represents the predicted four distance, $\boldsymbol{d}_i^*$ represents the distance from position to four sides of ground-truth, $\theta_i$ is the predicted rotated angle of the object, $\theta_i^*$ represents the rotated angle of ground-truth, $\lambda_d$ and $\lambda_a$ are the weights to balance the importance among three losses. The loss of classification $L_{cls}$ is the Focal loss [5], the distance loss $L_d$ is defined by IoU loss in UnitBox [17], the loss of rotated angle $L_a$ is the same in EAST [18].

*C. Locality-Aware Score Alignment*

We note that location regression and classification are independent for the one-stage detection method, and they may lead to the problem that the predicted bounding boxes with high scores have a low overlap rate with ground-truth. Furthermore, our method adds a new independent branch to regress angles, which aggravates this problem. Two-stage detection methods re-recognize the proposal regions by the detection stage, but the process of re-recognition increases computational and time costs as well. We propose the locality-aware score alignment by sampling several positions inside the bounding box from the score map, and the flowchart as shown in Fig. 5. We design several sampling point distributions relative to the rotated bounding box in Fig. 6. We compute the exact scores of the score map at sampling points by bilinear interpolation to overcome the misalignment caused by location quantization. The final score is the average of these computed scores at sampling points.

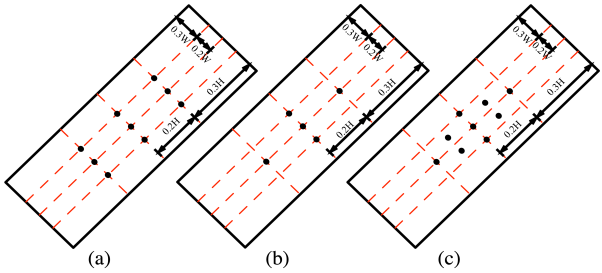

Fig. 6. Sampling point distributions of the locality-aware score alignment. (a) 9-point rectangular distribution. (b) 5-point diamond distribution. (c) 9-point diamond distribution.

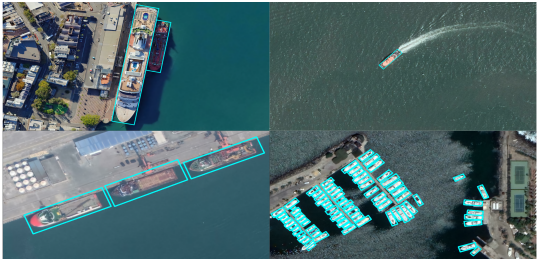

Fig. 7. Samples of annotated images in HRSD.

## III. EXPERIMENTAL RESULTS

Our experiments are undertaken on HRSC2016 and our dataset HRSD to evaluate and validate the performance of our algorithm. The experience settings and analysis of the experiment results are described as follows.

### A. Dataset Setup

Our HRSD dataset is collected at shoreside, sea and harbor, which are obtained from Google Earth with several resolutions to ensure the diversity of imagery background, as shown in Fig. 7. Considering the dataset bias, we selected ships with different categories and scales as detected targets. In total, we collected 2499 images with 9269 instances. The image sizes range from 600x300 to 7000x3500. According to the conventions of deep learning dataset, we made three splits to HRSD: training set, validation set, and test set. The training set contains 1249 images and 4431 instances, the validation set contains 424 images and 1712 instances and the test set contains 824 images and 3123 instances.

### B. Experiment Design and Implementation Details

We use the ResNet-101 pre-trained on ImageNet to initialize our backbone. Data augmentation, including horizontal flipping, random rotation, and random cropping, is adopted to strengthen the training ability and improve the generalization of our model. We train on two GPUs for 90k iterations with stochastic gradient descent. The initial learning rate is 0.001 and is divided by ten for every 10k iterations. The momentum is 0.9 and weight decay is 0.0001. The size of input images is fixed as 512 x 512 due to the limitation of GPU memory. The average precision (AP) is used as metrics to evaluate the performance of each detector.

TABLE I
COMPARISONS WITH THE DETECTION METHODS

| Methods | Anchor-based regression | Multi-level prediction | AP (%) HRSC2016 | AP (%) RSSD |
|---|---|---|---|---|
| R2CNN [12] | √ | / | 75.7 | 74.3 |
| R-DFPN [13] | √ | √ | 79.6 | 80.2 |
| Rotated FCN [14] | / | / | 82.3 | 78.3 |
| RRD [15] | √ | / | 84.3 | 82.3 |
| RoI Transformer [16] | √ | / | 86.2 | 84.8 |
| LAFCR-11 | √ | √ | 90.3 | 88.3 |

TABLE II
RESULT OF LOCALITY-AWARE SCORE ALIGNMENT

| Methods | Distribution | Number of points | AP (%) HRSC2016 | AP (%) RSSD |
|---|---|---|---|---|
| Baseline | / | / | 83.3 | 81.2 |
| LARSD-1 | Rectangle | 9 | 85.5 | 83.3 |
| LARSD-2 | Diamond | 5 | 86.0 | 83.7 |
| LARSD-3 | Diamond | 9 | 87.2 | 85.3 |
| LARSD-4 | Diamond | 13 | 87.3 | 85.4 |

### C. Comparison to State-of-the-Art Ship Detection Methods

The performance of our proposed method is compared with five competitive methods: (1) rotated region-based CNN (R2CNN) [12], (2) rotation dense feature pyramid networks (R-DFPN) [13], (3) rotated ship detection based on fully convolutional network (Rotated FCN) [14], (4) rotation sensitive regression detector (RRD) [15], and (5) RoI Transformer [16]. Table I summarizes the experiment results of our method and these comparisons. Conducted both on HRSC2016 and our dataset, our method achieves state-of-the-art performance, 90.3% and 88.3% AP, respectively.

### D. Experiment Analysis

#### 1) Effect of Locality-Aware Score Alignment

We design two sampling point distributions: diamond and rectangle. LARSD-1 is 9-point rectangular distribution, as shown in Fig. 6 (a). LARSD-2, LARSD-3, and LARSD-4 are diamond distribution, and the numbers of sampling points are 5, 9, and 13, respectively. To better analyze the effect of each proposed strategy, we build a baseline, which only uses UNet-like multi-scale CNN, without LASA, multi-level prediction and anchor-based regression. The experience results in Table II show that locality-aware score alignment with different distributions has a positive effect on AP. Compared to LARSD-1 and LARSD-3, we find diamond distribution has better performance due to that the diamond arrangement is similar to the shape of ships. We notice that properly increasing the number of sampling points is a promising approach for higher performance. Notably, the superior number of sampling points is 9.

#### 2) Effect of Multi-Level Prediction

Multi-level prediction is to alleviate the difficulty of recognizing objects under vastly different scales. It can produce multi-scale feature representations without image pyramids. Table III shows the result of anchor-based regression. The baseline, LARSD-5, and LARAD-6 detect objects by the single feature map (*P2*, *P3*, and *P4*, separately). The AP significantly decreases with the size of feature maps shrinking, which indicates that the high-resolution features are crucial for detection. We adopt the multi-level prediction strategy in LARSD-7, which utilizes the feature maps {*P2, P3, P4*} to predict the location and confidence. Compare with the baseline, LARSD-7 shows a minor improvement in AP. However, we further compare the location results of the baseline and LARSD-7, as shown in Fig. 8, which can be observed that the multi-scale prediction has more accurate location than the single-scale one.

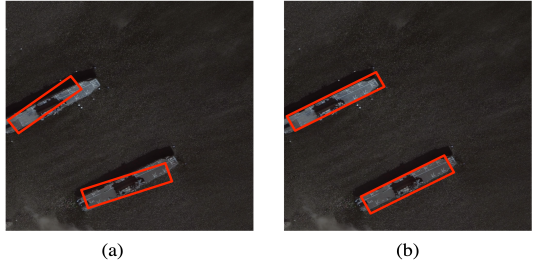

Fig. 8. Detection results with (a) baseline and (b) LARSD-7.

TABLE III
RESULT OF MULTI-LEVEL PREDICTION

| Methods | LASA | Level of prediction | AP (%) HRSC2016 | AP (%) RSSD |
|---|---|---|---|---|
| Baseline | / | P2 | 83.3 | 81.2 |
| LARSD-5 | / | P3 | 80.2 | 78.0 |
| LARSD-6 | / | P4 | 79.3 | 78.0 |
| LARSD-7 | / | P2+P3+P4 | 84.3 | 82.3 |
| LARSD-8 | √ | P2+P3+P4 | 88.0 | 86.1 |

TABLE IV
RESULT OF ANCHOR-BASED REGRESSION

| Methods | LASA | Level of prediction | Anchor-based regression | AP (%) HRSC2016 | AP (%) RSSD |
|---|---|---|---|---|---|
| Baseline | / | P2 | / | 83.3 | 81.2 |
| LARSD-9 | / | P2 | √ | 86.1 | 85.7 |
| LARSD-10 | / | P2+P3+P4 | √ | 86.9 | 86.3 |
| LARSD-11 | √ | P2+P3+P4 | √ | 90.3 | 88.3 |

*3) Effect of Anchor-Based Regression*

The advantage of using the anchor strategy can be interpreted as it reduces the difficulty of location regression, but it will also bring the challenge of classification accordingly. For ship detection, it is often intractable to acquire accurate location due to the variety of the scales and aspect ratios. Thus, we adopt the anchor-based distance regression to improve the performance of locations. As Table III and Table IV show, the methods with anchor-based regression, LARSD-9, LARSD-10, and LARSD-11, outperform the counterparts without anchor-based regression, the baseline, LARSD-7 and LARSD-8, and gain improvements of 4.5%, 4%, and 2.2% in AP, respectively.

## IV. CONCLUSION

In this paper, we propose a one-stage rational object detection method to detect rotated objects efficiently and accurately, especially for ships. Furthermore, we build a new rotated ship detection dataset and compare our method with other advanced rotated detection methods on HRSC2016 and our datasets. The experience shows that our model LARSD has state-of-the-art performance in ship detection. For the future work, we will extend our model application to the multi-class detection dataset and combine the study on semantic segmentation with our feature extract network to design a more effective network.